\documentclass[10pt,twocolumn,letterpaper]{article}

\usepackage{iccv}              

\definecolor{iccvblue}{rgb}{0.21,0.49,0.74}
\usepackage[pagebackref,breaklinks,colorlinks,allcolors=iccvblue]{hyperref}

\usepackage{textcomp, gensymb}

\usepackage{booktabs}
\usepackage{amsmath}
\usepackage{multirow}
\usepackage{graphicx}
\usepackage{colortbl}
\usepackage{amssymb}
\usepackage{overpic}

\usepackage{soul}
\graphicspath{{figures/}}

\usepackage[capitalize]{cleveref}
\crefname{section}{Sec.}{Secs.}
\Crefname{section}{Section}{Sections}
\Crefname{table}{Table}{Tables}
\crefname{table}{Tab.}{Tabs.}

\def\paperID{8172} 
\def\confName{ICCV}
\def\confYear{2025}

\title{VA-MoE: Variables-Adaptive Mixture of Experts for \\Incremental Weather Forecasting \includegraphics[scale=0.06]{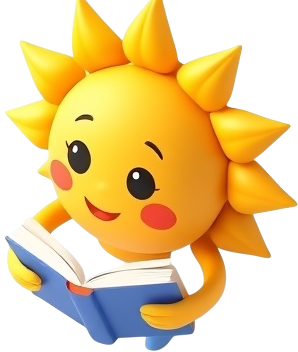} }

\author{%
 Hao Chen$^{1\dagger}$ \;\;     
 Han Tao$^{1\dagger}$\;\; 
 Guo Song$^{1\ast}$  \;\; 
 Jie Zhang$^{1}$\;\; 
 Yonghan Dong$^{2}$\;\;    
  Yunlong Yu$^{3}$ \;\;
 Lei Bai$^{4}$ \\  
 $^1$Hong Kong University of Science and Technology (HKUST)\;\;
 $^2$Huawei Technologies Ltd.  \\
  $^3$Zhejiang University  \;\; 
  $^4$Shanghai AI Laboratory\\
  {\tt\small $\dagger$Equal contribution }
  {\tt\small $^\ast$Corresponding author  }
 {\tt\small \{hchener, thanad\}@connect.ust.hk } \\
 {\tt\small \{songguo, csejzhang\}@ust.hk }
 {\tt\small dongyonghan@huawei.com }
 {\tt\small yuyunlong@zju.edu.cn }
 {\tt\small bailei@pjlab.org.cn }
}

\begin{document}
\maketitle

\begin{abstract}

This paper presents Variables-Adaptive Mixture of Experts (VA-MoE), a novel framework for incremental weather forecasting that dynamically adapts to evolving spatiotemporal patterns in real-time data. Traditional weather prediction models often struggle with exorbitant computational expenditure and the need to continuously update forecasts as new observations arrive. VA-MoE addresses these challenges by leveraging a hybrid architecture of experts, where each expert specializes in capturing distinct sub-patterns of atmospheric variables (e.g., temperature, humidity, wind speed). Moreover, the proposed method employs a variable-adaptive gating mechanism to dynamically select and combine relevant experts based on the input context, enabling efficient knowledge distillation and parameter sharing. This design significantly reduces computational overhead while maintaining high forecast accuracy. Experiments on ERA5 dataset demonstrate that VA-MoE performs comparable against state-of-the-art models in both short-term (e.g., 1–3 days) and long-term (e.g., 5 days) forecasting tasks, with only about 25\% of trainable parameters and 50\% of the initial training data. Code: \url{https://github.com/chenhao-zju/VAMoE}

\end{abstract}

\section{Introduction}
\label{sec:intro}

Weather forecasting remains one of the most critical scientific challenges with profound socio-economic impacts. While traditional Numerical Weather Prediction (NWP) \cite{nwp} relies on solving complex partial different equations to simulate atmospheric dynamics, recent advances in data-driven artificial intelligence \cite{fourcastnet,panguweather,fengwu, aurora} have revolutionized earth system modeling. Deep learning models, in particular, have demonstrated remarkable potential in capturing diverse spatiotemporal patterns across weather and climate phenomena, giving rise to the emerging field of AI for Weather.

However, a fundamental limitation of existing AI4Weather approaches lies in their assumption that all variables are available synchronously during training and inference. In reality, meteorological variables are often heterogeneous in terms of data sources, collection frequencies, and spatial-temporal distributions. For instance, the upper-air variables (e.g., temperature profiles) are sparse and sampled via radiosondes/satellites, while the surface variable (e.g., precipitation, wind) are dense but updated in near-real-time. This asynchrony poses significant challenges: when introducing new variables (e.g., satellite-derived aerosol data), existing models must be entirely retrained from scratch, incurring prohibitive computational costs. For example, Pangu \cite{panguweather} required 64 days and 192 V100 GPUs for a full retraining cycle.




\begin{figure}[t]
    \centering
    \begin{overpic}[width=\linewidth]{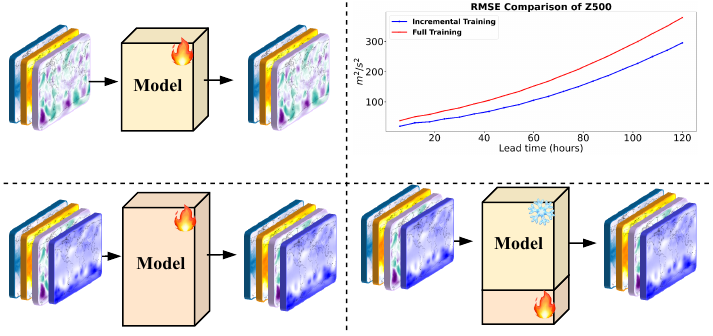} 
    \put(15,22.5){\scriptsize (a) Pre-Training}
    \put(65,22.5){\scriptsize (b) Comparison}
    \put(15,-3){\scriptsize (c) Full-Training}
    \put(60,-3){\scriptsize (d) Incremental-Training}
    \end{overpic}
    \vspace{0.01mm}
    \caption{Illustration of two different training paradigms of weather forecasting when new variables arrive: (c) Full-Train the whole model (d) Incremental-Train only the module corresponding new variables. (b) Performances of these two paradigms. }
    \label{fig:task}
\end{figure}

To address this challenge, we propose Incremental Weather Forecasting (IWF), a novel paradigm designed to dynamically expand meteorological forecast variables in response to increasing observational data. By enabling parameter-efficient variable scaling, IWF revolutionizes operational meteorology by offering a streamlined alternative to conventional models. This innovation tackles a critical limitation in the field: the need for scalable modeling solutions as datasets grow gradually, ensuring adaptability without compromising computational efficiency.

A pivotal limitation of existing incremental approaches is catastrophic forgetting, i.e., when introducing new variables, pre-trained parameters drift towards new distributions, causing significant performance degradation on original variables. To overcome this, we present the \textbf{Variables-Adaptive Mixture of Experts (VA-MoE)} model, the first expert-based system explicitly designed for hierarchical atmospheric variables.

Mixture of Experts (MoE) models inherently enable scalable representation learning for multi-variable atmospheric datasets by dynamically expanding expert modules as input dimensions grow. While traditional MoE architectures excel in representational capacity, their parallelized design which distributes uniform weights across experts, often leads to homogeneous outputs, limiting specialized knowledge acquisition. To address this limitation, we introduce VA-MoE, an auxiliary-loss-free framework that decouples expert specialization from computational efficiency. VA-MoE employs a phased training strategy: Initial experts are pretrained on base variables, then incrementally expanded to accommodate new variables. Crucially, the pretrained experts are frozen during expansion phases to prevent catastrophic forgetting, ensuring robust retention of prior expertise while adapting to novel inputs. To foster expert diversity without auxiliary losses, VA-MoE incorporates variable index embeddings, i.e., positional metadata encodings that guide experts to develop domain-specific specialization. These embeddings dynamically activate contextually relevant computational pathways for each variable, optimizing resource allocation at inference time. By integrating variable index embeddings that encode spatial-temporal metadata, VA-MoE directs experts to specialize in specific atmospheric domains. This context-aware routing reduces inference latency compared to static MoE models while maintaining expert diversity.

Besides, we propose a gradient-scaled variable-adaptive loss function that dynamically aligns variable optimization rates with their inherent spatiotemporal characteristics. By quantifying the distinct temporal evolution patterns of atmospheric variables, our method allocates differential loss weights, i.e., larger gradient magnitudes prioritize fast-transient fields like temperature to capture immediate atmospheric dynamics, while gradual weight adjustments stabilize slow-changing fields such as geopotential height to preserve long-term system behavior.

As shown in \cref{fig:task}, our paradigm employs a two-phase training paradigm: an initial phase optimizing base atmospheric variables, followed by an incremental phase integrating new variables while freezing previously trained experts. Experiments with the same 40-year dataset reveal that our paradigm \cref{fig:task}(a)+(d) achieves superior z500 predictions performance compared to full retraining \cref{fig:task}(a)+(c).
 

In conclusion, the contributions of this work include:
\begin{itemize}
    \item This work initiates systematic research on incremental learning paradigms for weather forecasting. We propose the quantitative benchmark to evaluate the trade-offs between model scalability and generalization in incremental weather modeling. 
    
    \item We present Variables-Adaptive Mixture of Experts (VA-MoE), the first framework tailored for incremental atmospheric modeling. VA-MoE achieves expert specialization through contextual variable activation driven by variable index embeddings, enabling dynamic assignment of experts to variables during both training and inference. 

    \item Extensive experiments on the ERA5 dataset demonstrate that VA-MoE achieves comparable performance for surface variables, while delivering superior accuracy in upper-air variables against the existing competitors under 50\% reduced dataset size and 25\% fewer parameters.
\end{itemize}

\section{Related work}
\label{sec:related}

\subsection{Data-Driven Weather Forecasting}

In recent decades, traditional NWP \cite{nwp} methods have dominated the weather forecasting field due to the robust predictions and rigorous mathematical validation. However, NWP methods \cite{molteni1996ecmwf, ritchie1995implementation} require training from scratch for new prediction, resulting in slow computation and high costs.


Recent advances in deep learning have catalyzed a paradigm shift toward data-driven models for medium-range weather forecasting. Pioneering works like FourCastNet \cite{fourcastnet} leverage Fourier Neural Operators (FNOs) \cite{li2020fno} to learn spatiotemporal weather patterns, while Pangu-Weather \cite{panguweather} employs a 3D vision transformer \cite{vit, liu2021swin} to model atmospheric dynamics. Subsequent innovations, including Neural Operators \cite{sfno, xiong2023koopmanlab, li2020fno} and specialized architectures \cite{graphcast, fengwu, fuxi, diffcast, nowcastnet, gencast, oneforecast, probabilistic}, have achieved accuracy rivaling NWP at a fraction of computation.

While existing data-driven models excel at fixed-variable forecasting after costly initial training, their inflexibility poses a critical barrier: retraining from scratch is required to incorporate new variables, rendering cross-domain adaptation impractical. In contrast, our work introduces incremental weather forecasting, a novel paradigm evolving spatiotemporal patterns in real-time data. By decoupling variable-specific expertise from shared dynamics, our approach supports dynamic expansion to new atmospheric parameters with minimal retraining overhead.

\begin{figure*}[t]
  \centering
   \includegraphics[width=0.95\linewidth]{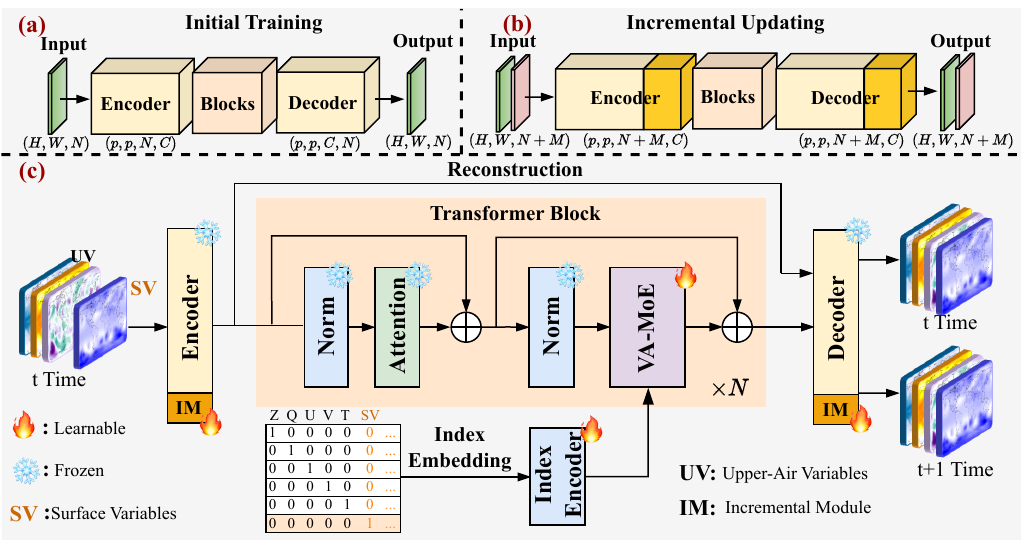}
   \caption{Illustration of the incremental weather forecasting paradigm. (a), (b) are the initial and incremental stages, respectively. (c) presents the detailed structure in both initial and incremental stages. During the initial stage, the model is trained with upper-air variables(UV). In the incremental stage, the \textcolor{red}{fire} components are trained while the \textcolor{blue}{snow} components remain frozen. The index embedding is trained by the Index Encoder and divided into six parts based on the upper-air variables (UV: Z, Q, U, V, T) and surface variables (SV). }
   \label{fig:framework}
   \vspace{-0.2cm}
\end{figure*}

\subsection{Mixture of Experts for Incremental Learning}

Advances in deep learning\cite{he2016deep, sppel} have spurred interest in parameter-efficient transfer strategies, where models adapt to new domains by fine-tuning only a subset of parameters. A key paradigm, incremental learning, addresses the stability-plasticity dilemma: retaining knowledge of prior tasks while integrating new concepts. However, real-world data often involves domain shifts, such as variations in variable distributions or task objectives, leading to two dominant research threads. Class Incremental Learning (CIL) \cite{goswami2024resurrecting, Kim_2024_CVPR, zhu2024rcl, tan2024semantically}, focuses on expanding classification tasks with new classes while Task Incremental Learning (TIL) \cite{ye2024online, yan2024orchestrate, seo2024learning, pan2024adaptive} adapts models to distinct task sequences. 

One of the most popular structures for addressing incremental learning tasks is the Mixture of Experts (MoE) model \cite{jacobs1991adaptive}, which learns representations of new concepts through different experts. Due to its sparse architecture, many works have adopted the MoE structure to reduce inference costs and enhance model capacity \cite{yang2024multi, wu2024omni}. An early work introducing MoE to incremental learning is Expert Gate \cite{expertgate}, which trains multiple backbones as different experts and assigns new domains to the relevant expert. Lifelong-MoE \cite{lifelongmoe} leverages pretrained experts and gates to retain the representational knowledge of the training domain. In addition to these approaches, MoE is used as an adapter in MoE-Adapter \cite{moeadapter}, where adapters are attached to the main structure. These works have demonstrated the promising performance of MoE in visual and natural language incremental learning.

In contrast to MoE designed for fixed-variable systems, our propose incremental learning of variable distributions by assigning distinct pressure-level groups to specialized experts. Unlike prior incremental learning literature that primarily focus on task or dataset-specific adaptation, our incremental paradigm targets variable-centric expansion, enabling models to adapt to evolving observational data without retraining from scratch.

\section{Methodology} \label{sec:method}


Our work focuses on a novel paradigm of incremental weather forecasting with VA-MoE. In this section, we first formally define the problem in \cref{subsec:problem}. Then, we present the structure of VA-MoE, consisting with the Channel-Adaptive Expert (CAE) and the Shared Expert in \cref{subsec:transformer}. The model during incremental stage is introduced in \cref{subsec:implementation}. Finally, loss function is presented in \cref{subsec:loss}.


\subsection{Overview} \label{subsec:problem}

For the weather forecasting task, the AI-based model $\Phi$ aims to predict the future weather variables $\text{X}^{t+1}$ with the historical weather variables $\text{X}^{t}$, i.e., $\text{X}^{t+1}=\Phi (\text{X}^{t})$.

In the incremental weather forecasting task, the weather variables $\text{X}^{t}$ are divided into two sets: (i) the initial training variables at $t$ time $\text{X}^{t}_{h} \in \mathbb{R}^{H \times W \times N}$, and, (ii) the incremental variables at $t$ time $\text{X}^{t}_{sv} \in \mathbb{R}^{H \times W \times M}$. Unlike traditional weather forecasting tasks, where models are trained with all variables available simultaneously, the incremental weather forecasting task sequentially incorporates dynamic variables as they become available over time. The incremental weather forecasting process is structured into a sequence of learning stages: an initial training stage followed by subsequent training stages.

In the initial training stage, the available variables of 40 years are used to train a weather model, i.e., $\text{X}^{t+1}_{h} = \Phi_h (\text{X}^{t}_{h})$, as shown in \cref{fig:framework}(a). In the incremental training stage, the half variables used at the initial stage and the surface variables at the incremental stage are incorporated together, to train the newly added modules for the surface variables while keeping the model parameters from the initial stage frozen, i.e., $\text{X}^{t+1} = \Phi (\text{X}^{t})$, where $\Phi$ consists of both the fixed parameters of the pre-trained model $\Phi_h$ and the trainable parameters of $\Phi_{sv}$, $\text{X}^{t}$ consists of all the available variables at time $t$, $\text{X}^{t+1}$ consists of the predicted variables at time $t+1$, as shown in \cref{fig:framework}(b). In this work, the upper-air variables are utilized during the initial training stage while the surface variables become available and are incorporated during the incremental training stage. The upper-air variables include five different types, Z, Q, U, V, and T, each defined across 13 different levels.

\begin{figure}[t]
  \centering
   \includegraphics[width=1\linewidth]{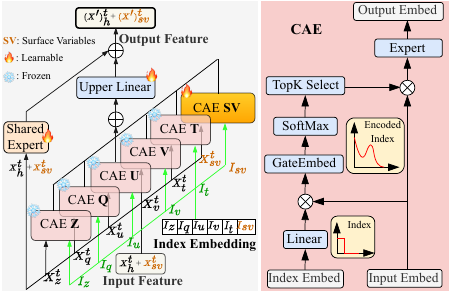}
   \caption{ 
   Illustration of the VA-MoE. In the \textbf{left subgraph}, the input features comprise both upper-air and surface variables. During the initial training stage, five distinct CAE modules process the upper-air variables. In the incremental stage, surface variables are handled by a dedicated module, $\mathrm{CAE_{SV}}$, while the original five CAEs remain frozen to preserve learned representations. The \textbf{right subgraph} details the CAE module. The Index Embedding is selected through a multi-layer process and applied to weight the Input Embedding via multiplication. Initially discrete, the index embedding is transformed into a continuous representation after passing through the GateEmbed layer, enabling it to dynamically reflect the relevance of specific channels to the input variables.}
   
   \label{fig:pccmoe}
\end{figure}

\subsection{VA-MoE} \label{subsec:transformer}


We introduce index embedding within transformer blocks, a novel mechanism to dynamically guide experts in learning hierarchical relationships between meteorological variables. As new variables are incrementally integrated, corresponding experts are added to the transformer architecture and optimized via index-based affinity assignments. Building on this framework, we adapt the Mixture-of-Experts (MoE) paradigm which enables task-specific specialization and flexible integration of new experts—into weather forecasting transformers.


The structure of the proposed VA-MoE is illustrated in \cref{fig:framework}(c). The input features, extracted by the encoder, first pass through a normalization layer and a self-attention layer. The output of the self-attention layer is then combined with the residual connection. Then, another normalization layer is applied, followed by a VA-MoE module, which is also equipped with a residual connection. This process can be formulated as:
\begin{eqnarray}\label{eq:global_mix}
\begin{aligned}
  \mathbf{X}_{mid} &= \mathbf{X}_{in} + \rm{SA}(\text{LayerNorm}(\mathbf{X}_{in})),\\
  \mathbf{X}_{out} &= \mathbf{X}_{mid} + \text{VA-MoE}(\text{LayerNorm}(\mathbf{X}_{mid})),
\end{aligned} 
\end{eqnarray}
where $\mathbf{X}_{in}$ and $\mathbf{X}_{out}$ are the input and the output of each Transformer Block, respectively. 

\subsubsection{Variables and Index Embedding}
As illustrated in \cref{fig:pccmoe}, during the training stage, the input feature $\mathbf{X}^{t}_{h} \in \mathbb{R}^{H \times W \times N}$ consists of five different types of variables: $\mathbf{X}_{Z}^{t}, \mathbf{X}_{Q}^{t}, \mathbf{X}_{U}^{t}, \mathbf{X}_{V}^{t}, \mathbf{X}_{T}^{t} \in \mathbb{R}^{H \times W \times (N/5)}$, where, $H \times W$ denotes the spatial resolution. These five types of variables are concatenated and processed by a shared expert and five distinct Channel-Adaptive Experts (CAEs).

In addition to these variables, the framework introduces an extra one-hot index embedding $\mathbf{I}_h \in \mathbb{R}^{5 \times N}$ to guide experts in learning variable affinity. The number $N$ corresponds to the number of variables, and 5 corresponds to the number of variable types. Similar to the input weather features, the index embedding is composed of five separate embeddings: $\mathbf{I}_{Z}, \mathbf{I}_{Q}, \mathbf{I}_{U}, \mathbf{I}_{V}, \mathbf{I}_{T} \in \mathbb{R}^{1\times 1\times N}$. This initial index embedding is then encoded into a latent space using a Linear layer, expressed as $\mathbf{I}_h=\mathrm{Linear}(\mathbf{I}_h)$.

\subsubsection{Channel-Adaptive Expert (CAE)} 

The key design of VA-MoE is the channel-adaptive expert, as shown in \cref{fig:pccmoe}. For each variable $Z$, the corresponding expert module processes both the feature embeddings of the input variables $\mathbf{X}^{t}_{h}$ and the index embedding $\mathbf{I}_{Z}$, which is formulated as: 
\begin{equation}
    \mathbf{X}^{t, CAE}_{Z} = \text{CAE}_{Z}(\mathbf{X}^{t}_{h}, \mathbf{I}_{Z}),
\end{equation}
where $\text{CAE}_Z$ is the abbreviation of channel-adaptive expert module for variable $Z$.

In the CAE module for variable $Z$, the index embedding $\mathbf{I}_{Z}$ is encoded through a linear layer for feature projection and then multiplied channel-wise with the input embedding $\mathbf{X}^{t}_{h}$. The resulting fused feature is processed by the GateEmbed layer, which produces a gate embedding to enhance the fusion of the index and variable embeddings. This enhanced feature is then normalized using a SoftMax layer and filtered through a TopK operation, which selects the top-K high-rank channels. This process yields the GateIndex $\mathrm{GI}_{Z} \in \mathbb{R}^{H \times W \times K}$ and GateWeight $\mathrm{GW}_{Z} \in \mathbb{R}^{H \times W \times K}$. The process is formulated as:
\begin{equation}
    \mathbf{I}_{Z}^{topk}, \mathbf{W}_{Z}^{topk} = \text{TOP}_{k}(\text{SoftMax}(\text{MLP}_Z(\mathbf{X}^{t}_{h} \odot \mathbf{I}_{Z}))),
\end{equation}
where $ \mathbf{I}_{Z}^{topk}$ and $ \mathbf{W}_{Z}^{topk}$ denote the index and weight of the Top-k features. Once the GateIndex and GateWeight are computed, the input embedding is selectively processed based on the GateIndex, while the variable embedding is weighted channel-wise according to the GateWeight. The GateEmbed layer uses variable embeddings to guide the index embedding, enabling the creation of a variable-specific confidence matrix. Within the CAE module, this embedding enhances the confidence matrix by incorporating both semantic and positional information, rather than relying solely on positional information.

The selected features for variable $Z$ are obtained with: 
\begin{equation}
    \mathbf{X}^{t, selected}_{Z} = \mathbf{W}_{Z}^{topk} * \text{Select}_{Z}(\mathbf{X}^{t}_{h}, \mathbf{I}_{Z}^{topk}),
\end{equation}
where $\text{Select}_{Z}$ denotes selecting Top-K embedding from the input feature $\mathbf{X}^{t}_{h}$ within the $\text{CAE}_{Z}$ module.



The selected and weighted embedding captures the key features associated with the specific type of variables. In the $\text{CAE}_{Z}$ module, these features serve as the training set to model the distribution of the variable $Z$ with $\text{Expert}_{Z}$. The expert module adopts a parallel structure, splitting the channels into multiple parts and enhancing each part with a sparse MoE. This process is formulated as:
\begin{equation}
    \mathbf{X}^{t, CAE}_{Z} = \text{Expert}_{Z}(\mathbf{X}^{t, selected}_{Z}),
\end{equation}
where $\mathbf{Expert}_{Z}$ is implemented with a multiple layer.

\subsubsection{Shared Expert.} 

The shared expert processes the overall features of all variables and operates in parallel with the variable-specific modules. Each variable is handled by its corresponding CAE module, and their outputs are summarized to produce the fused features $\mathbf{X}^{t, fused}_{h}$ for all variables. Since the channel count of the selected variable embeddings is lower than that of the original embeddings, an additional up-channel linear layer, $\text{Linear}_{\text{up}}$, is introduced. The fused features and the output from the shared expert are then combined via pixel-wise addition. This process is expressed as:
\begin{equation}
    \mathbf{X}^{t, fused}_{h} = \text{CAE}_{Z}(\mathbf{X}^{t}_{h}, \mathbf{I}_{Z}) + ... + \text{CAE}_{T}(\mathbf{X}^{t}_{h}, \mathbf{I}_{T}),
\end{equation}
\vspace{-0.5cm}
\begin{equation}
    \mathbf{(X')}^{t}_{h} = \text{Expert}_{shared}(\mathbf{X}^{t}_{h}) + \text{Linear}_{up}(\mathbf{X}^{t, fused}_{h} ).
\end{equation}

\subsection{Incremental Learning Stage}  \label{subsec:implementation}

When new variables are incrementally introduced to the model, specialized experts are dynamically integrated into the transformer blocks to process these variables. Concurrently, both the encoder and decoder architectures undergo synchronized updates to maintain compatibility with the expanded variable set.

\subsubsection{Encoder and Decoder Modules.} 

When $K$ new variables are introduced incrementally, the encoder’s input dimension expands to $H \times W \times (N+M)$, necessitating an adjustment to the convolutional layer’s kernel from $(3, 3, N, C)$ to $(3, 3, N + M, C)$. To accommodate this, the layer’s parameters are partitioned into two components: The $(3, 3, N, C)$ segment retains pretrained weights from earlier stages, ensuring continuity. The $(3, 3, M, C)$ segment, corresponding to the new variables, is initialized randomly to learn emergent patterns.



The same strategy is applied to the position embedding and the Decoder module.

\subsubsection{VA-MoE Module.} During the incremental learning phase, $M$ new experts are dynamically integrated into the MoE layer of each transformer block to accommodate $M$ newly introduced variables. In this phase, only the new experts and the shared expert are actively trained, while all preexisting experts remain frozen to preserve previously learned knowledge. 

\subsubsection{Index Embedding Module.} When $M$ new variables become available during the incremental stage, the index embedding changes from $\mathbf{I}_h \in \mathbb{R}^{(N\times l)\times N}$ to $\mathbf{I}_h \in \mathbb{R}^{(N\times l+M\times r)\times (N+M)}$, where each $M$ new variables has $r$ levels. Additionally, the index encoder needs to be retrained.

\subsection{Objective Function}
\label{subsec:loss}

To train the model, we introduce a novel dynamic prediction loss combined with a reconstruction loss. 

\subsubsection{Dynamic Prediction Loss.} The prediction loss aims to minimize the difference between the model's predicted results and the actual future weather variables. Unlike existing weather forecasting methods that treat all the variables equally, we argue that different variables, such as temperature and surface pressure, follow distinct distributions. Thus, we introduce a novel dynamic prediction loss that dynamically assigns a weight to each channel of the input data. This loss is formulated as:
\begin{equation}
   {Obj}_{pred} = (\mathbf{\hat{X}}^{t+1}-\mathbf{X}^{t+1}) \odot (\mathbf{\hat{X}}^{t+1}-\mathbf{X}^{t+1}) / e^\mathbf{w} + \mathbf{w},
\end{equation}
where $\odot$ denotes Hadamard product, $e$ denotes the base of the natural logarithm, $\mathbf{X}^{t+1} \in \mathbb{R}^{H\times W\times C}$ , and $\mathbf{w} \in \mathbb{R}^{1\times 1\times C}$ is a learnable vector. With this loss function, the model could dynamically prioritize variables based on their distributional characteristics.


\begin{table}[t]
  \centering
  \small
  \resizebox{0.85\linewidth}{!}{
    \begin{tabular}{clc}
    \toprule
    \textit{Name} & \multicolumn{1}{c}{Description} & Levels \\
    \midrule
    \multicolumn{3}{c}{\textbf{Upper-Air}} \\
    \midrule
    \textit{Z} & \textit{Geopotential} & 13 \\
    \textit{Q} & \textit{Specific humidity} & 13 \\
    \textit{U} & \textit{x-direction wind} & 13 \\
    \textit{V} & \textit{y-direction wind} & 13 \\
    \textit{T} & \textit{Temperature} & 13 \\
    \midrule
    \multicolumn{3}{c}{\textbf{Surface-Incremental Learning}} \\
    \midrule
    \textit{u10} & \textit{x-direction wind at 10m height} & Single \\
    \textit{v10} & \textit{y-direction wind at 10m height} & Single \\
    \textit{t2m} & \textit{Temperature at 2m height} & Single \\
    \textit{msl} & \textit{Mean sea-level pressure} & Single \\
    \textit{sp} & \textit{Surface pressure} & Single \\
    \bottomrule
    \end{tabular}}%
  \caption{A summary of atmospheric variables. The 13 levels are 50, 100, 150, 200, 250, 300, 400, 500, 600, 700, 850, 925, 1000 hPa. `Single' denotes the variables under earth's surface. }
  \label{tab:vars}
\end{table}

\subsubsection{Reconstruction Loss.}
We introduce an additional reconstruction loss that connects the encoder and decoder directly, formulated as:
\begin{equation}
   {\rm{Obj}}_{recon} = (\mathbf{\hat{X}}^{t}-\mathbf{X}^{t})^{2},
\end{equation}
where $\mathbf{\hat{X}}^{t}=\text{Dec}(\text{Enc}(\mathbf{X}^{t}))$ represents the reconstructed variables. With the loss function, the encoder and decoder modules specialize in encoding and decoding features, while the intermediate transformer blocks focus on learning the data distribution. This ensures that the decoding process is entirely handled by decoder, allowing the intermediate blocks to focuse on their task and not participate in decoding, thereby optimizing network efficiency.

To this end, the model is trained with the final objective function, a combination of the dynamic prediction loss and reconstruction loss. The final objective is formulated as:
\begin{equation}
    {\rm{Obj}}_{final} = {\rm{Obj}}_{pred} + \lambda {\rm{Obj}}_{recon},
\end{equation}
where $\lambda$ denotes a hyper-parameter.

\section{Experiments}
\label{sec:experiment}

\textbf{Dataset.} In this work, we conduct experiments on a popular weather dataset, \ie, ERA5\cite{era5}, provided by the ECMWF. ERA5 dataset is a reanalysis atmospheric dataset, consisting of the atmospheric variables from 1979 to the present day with a 0.25\degree spatial resolution with $721 \times 1440$. We train the model on 40-year dataset in the initial stage and continuously train the model on 20-year dataset in the incremental stage, which contains the weather variables from 1979 to 2020 and 2000 to 2020 year, respectively. And we test the model on one-year dataset of weather variables in 2021. In this work, the model processes 5 upper-air variables and 5 surface variables as \cref{tab:vars}, where the upper-air variables are exploited in the training stage, and the surface variables are exploited in the incremental stage.

\textbf{Implementation Details.} The main structure of this work follows the novel backbones \cite{flashattention, pmnet, mcinet}. We apply the AdamW optimizer with 0.0002 and 0.00005 learning rates to initial and incremental stages, respectively. In two stages, we train 100 epochs and set batch size to 16. Our model is trained with PyTorch using 16 A100 GPUs. 

\begin{table*}[t]
  \centering
  \makebox[\textwidth]{
\resizebox{\linewidth}{!}{
    \begin{tabular}{l@{\hspace{0.05cm}}|c@{\hspace{0.2cm}}|c@{\hspace{0.2cm}}|c@{\hspace{0.2cm}}c@{\hspace{0.2cm}}c@{\hspace{0.05cm}}|c@{\hspace{0.2cm}}c@{\hspace{0.2cm}}c@{\hspace{0.05cm}}|c@{\hspace{0.2cm}}c@{\hspace{0.2cm}}c@{\hspace{0.05cm}}|c@{\hspace{0.2cm}}c@{\hspace{0.2cm}}c@{\hspace{0.05cm}}|c@{\hspace{0.2cm}}c@{\hspace{0.2cm}}c@{\hspace{0.05cm}}}
    \toprule
        & Dataset & Iteration  & \multicolumn{3}{c|}{T2M ($K$) $\downarrow$} & \multicolumn{3}{c|}{U10 ($m/s$) $\downarrow$} & \multicolumn{3}{c|}{V10 ($m/s$) $\downarrow$} & \multicolumn{3}{c|}{MSL ($Pa$) $\downarrow$} & \multicolumn{3}{c}{ SP ($Pa$) $\downarrow$} \\
           
        & (years) & ($\times 10^{4}$) & 6h & 72h & 120h & 6h & 72h & 120h & 6h & 72h & 120h & 6h & 72h & 120h & 6h & 72h & 120h \\
    \midrule
               \multicolumn{18}{c}{Plain Training} \\
    \midrule
    ViT$^*$ \cite{vit}  &  1979-2020  &  40   & 0.72  & 1.35  & 1.86  & 0.66  & 1.98  & 3.01  & 0.68  & 2.02  & 3.11  & 40.2  & 208.5  & 393.9  & 63.3  & 222.1  & 397.0  \\
    
    IFS \cite{ifs} &  1979-2020 &  40 & 1.09 & 1.38 & 1.74 & 0.96 & 1.87 & 2.78 & 0.99 & 1.93 & 2.87 & -  & -  & -  & -  & -  & - \\ 
    
    Pangu-Weather \cite{panguweather} &  1979-2020 &  40  & 0.82 & 1.09 & 1.53 & 0.77  & 1.63 & 2.54 & 0.79 & 1.68 & 2.65  & -  & -  & -  & -  & -  & - \\
    
    FourCastNet \cite{fourcastnet} &  1979-2020 &  40 & 0.82 & 1.02 & 1.77 & 0.82 & 2.08 & 3.34 & 0.84 & 2.11 & 3.41 & -  & -  & -  & -  & -  & - \\ 

    ClimaX \cite{climax} &  1979-2020 &  40 & 1.11 & 1.47 & 1.83 & 1.04 & 2.02 & 2.79 & -  & -  & - & -  & -  & -  & -  & -  & - \\ 

    Graphcast \cite{graphcast}  &  1979-2020 &  40  & \textbf{0.51} & \textbf{0.94} & \textbf{1.37} & \textbf{0.38} & 1.51 & 2.37 & -  & -  & - & 23.4  & 135.2  & 278.2  & -  & -  & - \\ 

    Fengwu \cite{fengwu}  &  1979-2020 &  40 & 0.58 & 1.03 & 1.41 & 0.42 & 1.53 & 2.32 & -  & -  & - & \textbf{23.2}  & 137.1  & 276.9  & -  & -  & - \\ 

    FuXi \cite{fuxi}  &  1979-2020  &  40  & 0.55 & 0.99 & 1.41 & 0.42 & 1.50 & 2.36 & \textbf{0.43}  & 1.54  & 2.44 & 27.2  & 136.7  & 282.9  & -  & -  & - \\ 

    \midrule

     VA-MoE &  1979-2020  &  40  & 0.57  & 1.03  & 1.42  & 0.43  & \textbf{1.41}  & \textbf{2.25}  & 0.44  & \textbf{1.46}  & \textbf{2.34}  & 27.5  & \textbf{131.1}  & \textbf{275.9}  & \textbf{57.1}  & \textbf{168.9}  & \textbf{302.4}  \\

    \midrule
    
           \multicolumn{18}{c}{Incremental Training from \textbf{65 Upper-Air Variables (79-20)} to \textbf{5 Surface Variables (79-20)}} \\
          
    \midrule

     VA-MoE (IL) &  1979-2020 &  20  & 0.58  & 1.05  & 1.45  & 0.48  & 1.47  & 2.33  & 0.47  & 1.54  & 2.41  & 27.9  & 137.3  & 281.6  & 59.3  & 173.4  & 312.4  \\
    
    \midrule
           \multicolumn{18}{c}{Incremental Training from \textbf{65 Upper-Air Variables (79-20)} to \textbf{5 Surface Variables (00-20)}} \\
          
    \midrule
    
      VA-MoE (IL) &  2000-2020 &  10  & 0.73  & 1.17  & 1.57  & 0.54  & 1.58  & 2.49  & 0.55  & 1.63  & 2.57  & 30.0  & 148.8  & 304.7  & 60.6  & 171.4  & 314.8  \\

    
    \bottomrule
    \end{tabular}
    }}
        \caption{Prediction performances on Incremental Training with 5 surface variables, i.e., T2M, U10, V10, MSL, and SP. * denotes running by ourselves. The best results are marked in \textbf{bold}. All experiments are in 0.25\degree with $721\times1440$ resolutions.} 
        \label{tab:cl70}
\end{table*}%

\begin{table*}[t]
    \centering

    \makebox[\textwidth]{
    \resizebox{\linewidth}{!}{
    \begin{tabular}{l@{\hspace{0.05cm}}|c@{\hspace{0.1cm}}|l@{\hspace{0.2cm}}l@{\hspace{0.2cm}}l@{\hspace{0.05cm}}|c@{\hspace{0.2cm}}c@{\hspace{0.2cm}}c@{\hspace{0.05cm}}|l@{\hspace{0.2cm}}l@{\hspace{0.2cm}}l@{\hspace{0.05cm}}|l@{\hspace{0.2cm}}l@{\hspace{0.2cm}}l@{\hspace{0.05cm}}|l@{\hspace{0.2cm}}l@{\hspace{0.2cm}}l@{\hspace{0.05cm}}}
    \toprule
        ~ & Para. & \multicolumn{3}{c|}{Z500($m^2/s^2$) $\downarrow$} & \multicolumn{3}{c|}{Q500($\times e^{-3}, g/kg) \downarrow$} & \multicolumn{3}{c|}{U500($m/s$) $\downarrow$} & \multicolumn{3}{c|}{V500($m/s$) $\downarrow$} & \multicolumn{3}{c}{T500($K$) $\downarrow$}  \\ 
        
        ~ &  (M) & \multicolumn{1}{c}{6h} & \multicolumn{1}{c}{72h} & \multicolumn{1}{c|}{120h} & \multicolumn{1}{c}{6h} & \multicolumn{1}{c}{72h} & \multicolumn{1}{c|}{120h} & \multicolumn{1}{c}{6h} & \multicolumn{1}{c}{72h} & \multicolumn{1}{c|}{120h} & \multicolumn{1}{c}{6h} & \multicolumn{1}{c}{72h} & \multicolumn{1}{c|}{120h} & \multicolumn{1}{c}{6h} & \multicolumn{1}{c}{72h} & \multicolumn{1}{c}{120h} \\ 

    \midrule

        IFS \cite{ifs} & - & 28.31 & 154.08 & 333.96 & 0.31 & 0.61 & 0.75 & 1.43 & 3.23 & 5.12 & 1.40 & 3.58 & 5.64 & 0.36 & 0.98 & 1.70 \\ 
        Pangu-Weather \cite{panguweather} & - & 24.88 & 167.90 & 391.26 & 0.25 & 0.55 & 0.69 & 0.96 & 3.13 & 4.73 & 0.91 & 3.52 & 5.15 & 0.27 & 0.94 & 1.56 \\ 
        Graphcast \cite{graphcast} & - & 15.23 & 125.42 & 275.35 & - & - & - & 0.77 & 2.86 & 4.49 & 0.74 & 2.92 & 4.67 & 0.23 & 0.87 & 1.48 \\ 
   
    \midrule

     VA-MoE &  665 & 19.28 & 134.63 & 295.52 & 0.17 & 0.49 & 0.62 & 0.84 & 2.99 & 4.71 & 0.84 & 3.04 & 4.89 & 0.25 & 0.76 & 1.36 \\ 

    \midrule
           \multicolumn{17}{c}{Incremental Training from \textbf{65 Upper-Air Variables (79-20)} to \textbf{5 Surface Variables (00-20)}} \\
          
    \midrule

    VA-MoE (IL) &  137 & 18.23 & 133.14 & 292.63 & 0.17 & 0.49 & 0.61 & 0.84 & 3.01 & 4.74 & 0.84 & 3.51 & 4.93 & 0.25 & 0.76 & 1.37 \\ 

    \bottomrule
    \end{tabular}
    }}
        \caption{Prediction performances on Initial Training with 5 upper-air variables. All experiments are in 0.25\degree with $721\times1440$ resolutions.} 
        \label{tab:pl}
\end{table*}

\begin{table*}[t]
    \centering
    \makebox[\textwidth]{
    \resizebox{\linewidth}{!}{
  
    \begin{tabular}{l@{\hspace{0.05cm}}|c@{\hspace{0.1cm}}|l@{\hspace{0.2cm}}l@{\hspace{0.2cm}}l@{\hspace{0.05cm}}|c@{\hspace{0.2cm}}c@{\hspace{0.2cm}}c@{\hspace{0.05cm}}|l@{\hspace{0.2cm}}l@{\hspace{0.2cm}}l@{\hspace{0.05cm}}|l@{\hspace{0.2cm}}l@{\hspace{0.2cm}}l@{\hspace{0.05cm}}|l@{\hspace{0.2cm}}l@{\hspace{0.2cm}}l@{\hspace{0.05cm}}}
    \toprule
        ~ & Para. & \multicolumn{3}{c|}{Z500 ($m^2/s^2$) $\downarrow$} & \multicolumn{3}{c|}{Q500 ($\times e^{-3}, g/kg) \downarrow$} & \multicolumn{3}{c|}{U500 ($m/s$) $\downarrow$} & \multicolumn{3}{c|}{V500 ($m/s$) $\downarrow$} & \multicolumn{3}{c}{T500 ($K$) $\downarrow$}  \\ 
        
        ~ &  (M) & \multicolumn{1}{c}{6h} & \multicolumn{1}{c}{72h} & \multicolumn{1}{c|}{120h} & \multicolumn{1}{c}{6h} & \multicolumn{1}{c}{72h} & \multicolumn{1}{c|}{120h} & \multicolumn{1}{c}{6h} & \multicolumn{1}{c}{72h} & \multicolumn{1}{c|}{120h} & \multicolumn{1}{c}{6h} & \multicolumn{1}{c}{72h} & \multicolumn{1}{c|}{120h} & \multicolumn{1}{c}{6h} & \multicolumn{1}{c}{72h} & \multicolumn{1}{c}{120h} \\ 

    \midrule

        ViT$^*$\cite{vit} & 307 & 33.38 & 209.4 & 517.81 & 0.22 & 0.61 & 1.06 & 1.24 & 3.66 & 6.52 & 1.22 & 3.76 & 7.41 & 0.42 & 1.18 & 2.40 \\ 
        ViT+MoE (light)$^*$\cite{dit-moe} & 609 & 37.92 & 207.11 & 405.73 & 0.22 & 0.60 & 0.78 & 1.30 & 3.84 & 5.87 & 1.27 & 3.89 & 6.11 & 0.46 & 1.23 & 2.02 \\ 
        ViT+MoE$^*$\cite{dit-moe} & 1113 & 28.31 & 169.61 & 356.02 & 0.23 & 0.56 & 0.72 & 1.21 & 3.46 & 5.44 & 1.23 & 3.54 & 5.69 & 0.35 & 1.07 & 1.83 \\ 

    \midrule
    
     VA-MoE &  665 & 20.59 & 139.02 & 302.13 & 0.18 & 0.49 & 0.62 & 0.91 & 3.02 & 4.76 & 0.91 & 3.08 & 4.97 & 0.27 & 0.92 & 1.59 \\ 
     
     VA-MoE (IL) &  137 & 20.29 & 138.52 & 301.41 & 0.18 & 0.50 & 0.63 & 0.91 & 3.04 & 4.79 & 0.91 & 3.10 & 5.03 & 0.27 & 0.93 & 1.60 \\ 

    \bottomrule
    \end{tabular}
    }}%
        \caption{ Architectural impact on 5 upper-air variables under 500 hPa. All experiments are in 1.5\degree with $128\times256$ resolutions.}  
        \label{tab:ablation}
    \vspace{-0.4cm}
        
\end{table*}


\begin{figure}[t]
  \centering
   \includegraphics[width=1.0\linewidth]{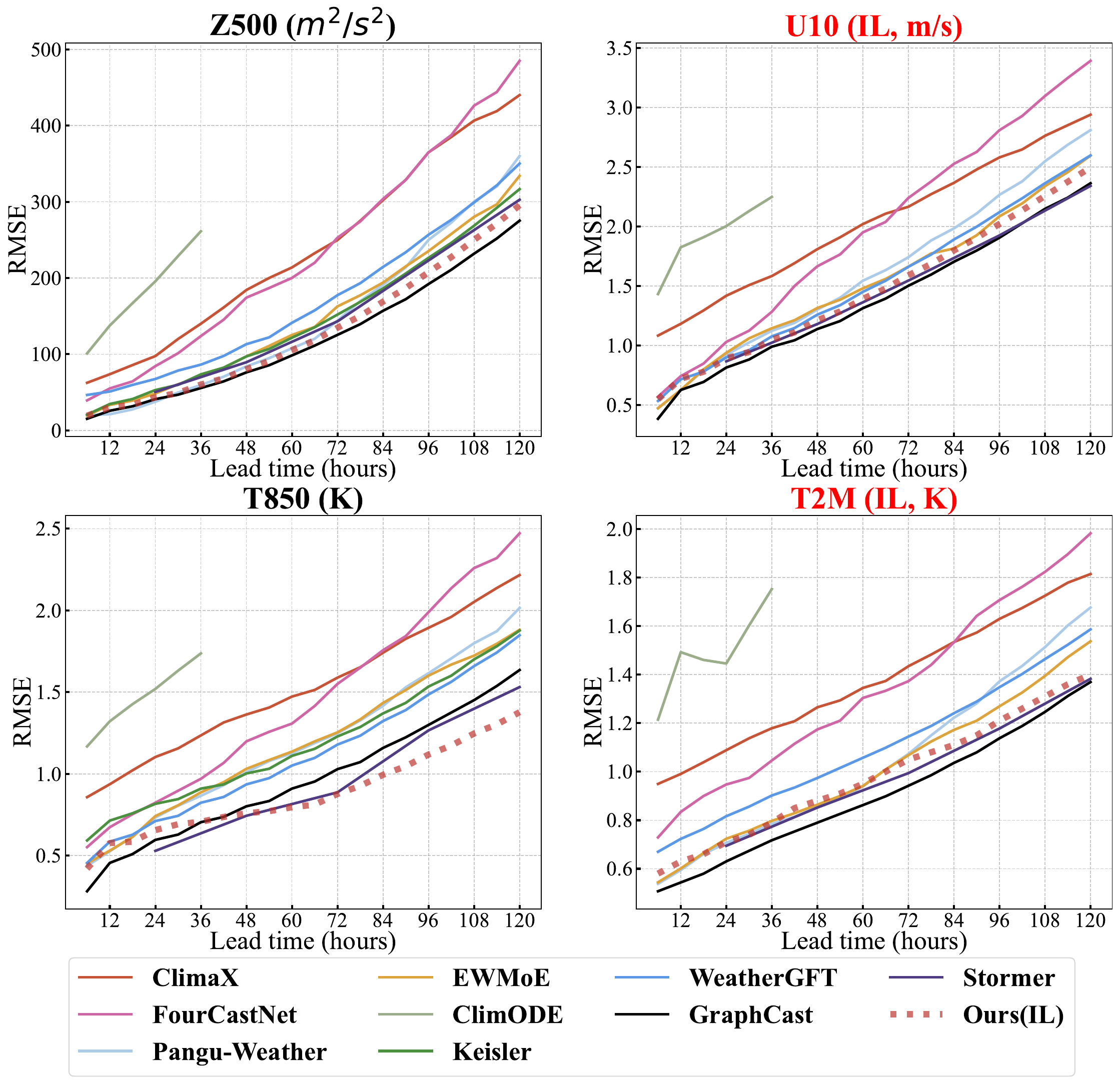}
   \caption{ 
   Comparative analysis of \textbf{RMSE} \boldmath{$\downarrow$} across 10 data-driven models for four variables, including Z500 and T850 (upper-air variables) in the initial stage, as well as T2M and U10 (incremental surface variables) in the incremental stage.
   } 
   \label{fig:sota}
\end{figure}

\subsection{Main Results}

\cref{fig:sota} presents a quantitative comparison of our structure with existing state-of-the-art approaches, including ClimaX \cite{climax}, Pangu-Weather \cite{panguweather}, ClimODE \cite{climode}, WeatherGFT \cite{weathergft}, FourCastNet \cite{fourcastnet}, EWMoE \cite{ewmoe}, Keisler \cite{keisler}, GraphCast \cite{graphcast}, and Stormer \cite{stormer}, for the 5-day weather prediction task. All the competitors are trained with both upper-air and surface variables on a $0.25^\circ$ weather dataset with a resolution of $721 \times 1440$.

For the surface variables, including T2M and U10, our VA-MoE demonstrates performance similar to the leading methods Stormer \cite{stormer} and Graphcast \cite{graphcast}. Compared to other competitors, our structure shows significant advantages in both short-term and long-term predictions. For the upper-air variable Z500, our structure delivers one of the best performances among all 10 methods, outperforming Pangu-Weather \cite{panguweather} and Stormer \cite{stormer}, and significantly surpassing ClimaX \cite{climax} and FourCastNet \cite{fourcastnet}, while slightly inferior to GraphCast \cite{graphcast}. For T850, our structure is slightly inferior to EWMoE \cite{ewmoe}, GraphCast \cite{graphcast}, and Stormer \cite{stormer} in the short-term predictions within 48 hours. After that period, our structure achieves the best performance in the long-term predictions.

\begin{figure*}[t]
  \centering
   \includegraphics[width=0.95\linewidth]{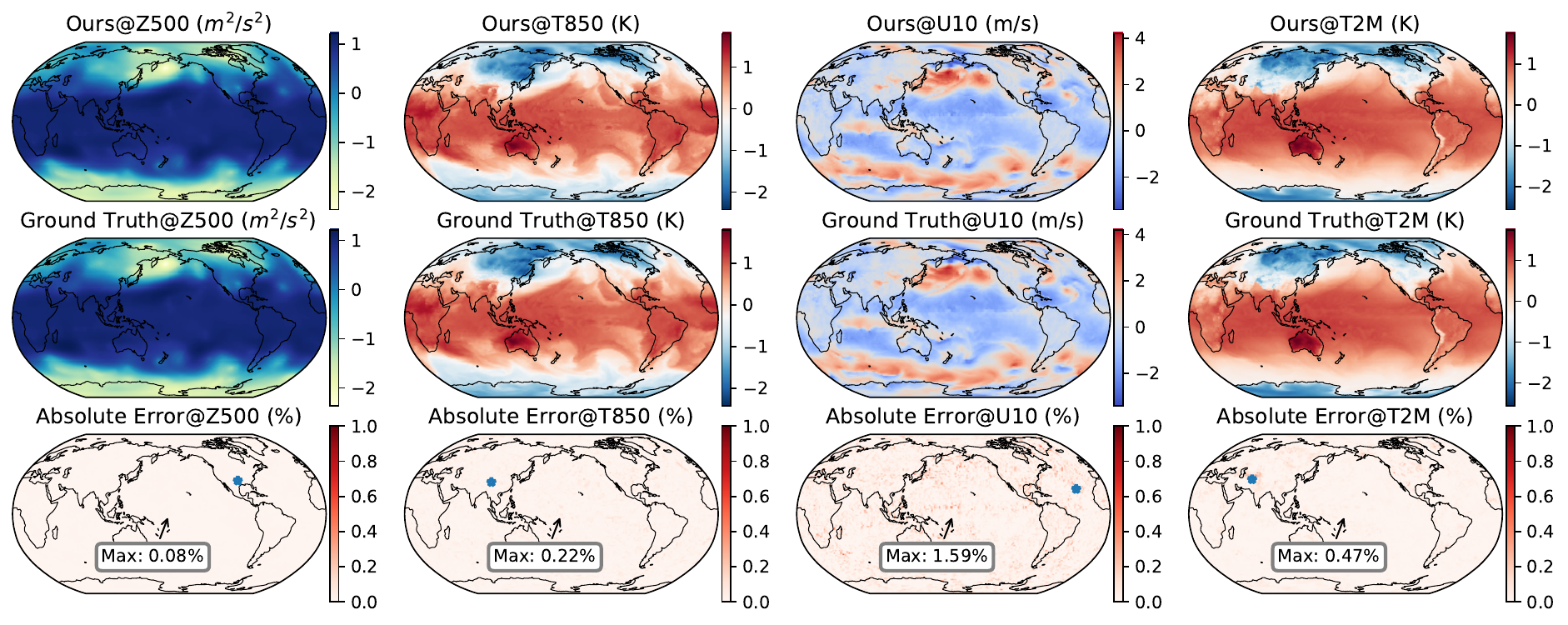}
   \vspace{-0.2cm}
   \caption{6-hour global weather prediction of upper-air and surface variables forecast visualization generated by VA-MoE framework.} 
   \label{fig:predict}
   \vspace{-0.4cm}
\end{figure*}

\textbf{Results of Surface Variables Prediction.} This study evaluates model performance on five surface variables: T2M, U10, V10, MSL, and SP. While competing structures are trained with both upper-air and surface variables, VA-MoE (IL) employs a two-stage training approach: initial training with upper-air variables followed by incremental training with all five surface variables.

Analysis of the results from \cref{tab:cl70} demonstrates that VA-MoE outperforms competing models across multiple variables and settings, particularly in U10 and V10. Although VA-MoE shows marginally weaker performance in T2M compared to GraphCast \cite{graphcast} and comparable results to FengWu \cite{fengwu} and FuXi \cite{fuxi}, it achieves superior long-term predictions for U10, V10, MSL, and SP. These indicate that while VA-MoE exhibits slight limitations in short-term forecasting, it excels in long-term prediction tasks.

When implementing the incremental training strategy, VA-MoE (IL) trained on a 40-year dataset with only half the training iterations achieves performance comparable to standard VA-MoE. Even with a reduced 20-year dataset, VA-MoE(IL) maintains acceptable performance using just a quarter of the iterations required by the standard approach. These results underscore the efficacy of incremental learning for atmospheric datasets and suggest its potential to replace conventional training methods in operational settings.

\textbf{Results of Upper-Air Variables Prediction.} This study evaluates model performance on upper-air variables, comparing three training approaches: (1) VA-MoE, trained exclusively on upper-air variables; (2) VA-MoE (IL), which incorporates incremental training on surface variables after initial upper-air training; and (3) competing models trained simultaneously on both upper-air and surface variables.

As shown in \cref{tab:pl}, VA-MoE achieves performance comparable to GraphCast \cite{graphcast} and outperforms IFS \cite{ifs} and Pangu-Weather \cite{panguweather} across both short- and long-term prediction. Notably, VA-MoE (IL) demonstrates robust predictive capabilities for upper-air variables even after incremental training on surface variables, despite using only a 20-year dataset - half the size of the initial training data. Furthermore, VA-MoE (IL) exhibits marginal performance improvements over the baseline in long-term predictions of Z500, confirming the absence of catastrophic forgetting during incremental training. These results highlight the efficacy of incremental training in VA-MoE(IL) for maintaining and enhancing predictive accuracy on original variables.

\subsection{Ablation Study}

\textbf{Impacts of Architecture.} As shown in \cref{tab:ablation}, we compare VA-MoE with other architectures, including ViT and ViT+MoE. VA-MoE significantly outperforms both ViT and ViT+MoE architectures across 6-hour, 72-hour, and 120-hour prediction horizons, despite ViT+MoE containing nearly twice the number of parameters. This performance gap highlights VA-MoE’s suitability for weather forecasting tasks as its channel-adaptive expert design ensures parameter efficiency while maintaining accuracy, particularly in incremental learning. Considering the computational complexity, \textbf{all ablation studies are in 1.5\degree with $128\times256$. }



\subsection{Visualization}


In the work, we visualize some predicted results and parameters. In \cref{fig:predict}, we visualize the 6-hour predicted results of Z500, T850, U10, and T2M. There are max 0.08\%, 0.22\%, 1.59\%, and 0.47\% absolute errors across 4 variables, including 2 upper-air variables and 2 surface variables. 

\section{Conclusion}
\label{sec:conclusion}

In this work, we proposed incremental weather forecasting, a novel task that addresses the challenge of dynamically expanding weather models to incorporate new variables without retraining from scratch. Our solution, Variables-Adaptive Mixture-of-Experts, enables parameter-efficient adaptation by selectively fine-tuning experts assigned to new variables through index embedding guidance, while preserving pretrained knowledge in existing parameters. Experimental results demonstrate that our approach achieves forecast accuracy comparable to fully retrained models while drastically reducing computational costs.

\section{Acknowledgements}

This research was supported by the Hong Kong RGC General Research Fund (152169/22E, 152228/23E, 162161/24E), Research Impact Fund (No. R5060-19, No. R5011-23), Collaborative Research Fund (No. C1042-23GF), NSFC/RGC Collaborative Research Scheme (No. 62461160332 \& CRS\_HKUST602/24), Areas of Excellence Scheme (AoE/E-601/22-R), the InnoHK (HKGAI), and Key R\&D  Program of Zhejiang Province 2025C01075.

{
    \small
    \bibliographystyle{ieeenat_fullname}
    \bibliography{main}

\begin{thebibliography}{47}
\providecommand{\natexlab}[1]{#1}
\providecommand{\url}[1]{\texttt{#1}}
\expandafter\ifx\csname urlstyle\endcsname\relax
  \providecommand{\doi}[1]{doi: #1}\else
  \providecommand{\doi}{doi: \begingroup \urlstyle{rm}\Url}\fi

\bibitem[Aljundi et~al.(2017)Aljundi, Chakravarty, and Tuytelaars]{expertgate}
Rahaf Aljundi, Punarjay Chakravarty, and Tinne Tuytelaars.
\newblock Expert gate: Lifelong learning with a network of experts.
\newblock In \emph{Proceedings of the IEEE Conference on Computer Vision and Pattern Recognition}, pages 3366--3375, 2017.

\bibitem[Bauer et~al.(2015)Bauer, Thorpe, and Brunet]{nwp}
Peter Bauer, Alan Thorpe, and Gilbert Brunet.
\newblock The quiet revolution of numerical weather prediction.
\newblock \emph{Nature}, 525\penalty0 (7567):\penalty0 47--55, 2015.

\bibitem[Bi et~al.(2023)Bi, Xie, Zhang, Chen, Gu, and Tian]{panguweather}
Kaifeng Bi, Lingxi Xie, Hengheng Zhang, Xin Chen, Xiaotao Gu, and Qi Tian.
\newblock Accurate medium-range global weather forecasting with 3d neural networks.
\newblock \emph{Nature}, 619\penalty0 (7970):\penalty0 533--538, 2023.

\bibitem[Bodnar et~al.(2024)Bodnar, Bruinsma, Lucic, Stanley, Brandstetter, Garvan, Riechert, Weyn, Dong, Vaughan, et~al.]{aurora}
Cristian Bodnar, Wessel~P Bruinsma, Ana Lucic, Megan Stanley, Johannes Brandstetter, Patrick Garvan, Maik Riechert, Jonathan Weyn, Haiyu Dong, Anna Vaughan, et~al.
\newblock Aurora: A foundation model of the atmosphere.
\newblock \emph{arXiv:2405.13063}, 2024.

\bibitem[Bonev et~al.(2023)Bonev, Kurth, Hundt, Pathak, Baust, Kashinath, and Anandkumar]{sfno}
Boris Bonev, Thorsten Kurth, Christian Hundt, Jaideep Pathak, Maximilian Baust, Karthik Kashinath, and Anima Anandkumar.
\newblock Spherical fourier neural operators: learning stable dynamics on the sphere.
\newblock In \emph{Proceedings of International Conference on Machine Learning}, 2023.

\bibitem[Chen et~al.(2023{\natexlab{a}})Chen, Dong, Lu, Yu, and Han]{sppel}
Hao Chen, Yonghan Dong, Zhe-Ming Lu, Yunlong Yu, and Jungong Han.
\newblock Self-prompting perceptual edge learning for dense prediction.
\newblock \emph{IEEE Transactions on Circuits and Systems for Video Technology}, 34\penalty0 (6):\penalty0 4528--4541, 2023{\natexlab{a}}.

\bibitem[Chen et~al.(2024{\natexlab{a}})Chen, Dong, Lu, Yu, and Han]{pmnet}
Hao Chen, Yonghan Dong, Zheming Lu, Yunlong Yu, and Jungong Han.
\newblock Pixel matching network for cross-domain few-shot segmentation.
\newblock In \emph{Proceedings of the IEEE Winter Conference on Applications of Computer Vision}, pages 978--987, 2024{\natexlab{a}}.

\bibitem[Chen et~al.(2024{\natexlab{b}})Chen, Yu, Dong, Lu, Li, and Zhang]{mcinet}
Hao Chen, Yunlong Yu, Yonghan Dong, Zheming Lu, Yingming Li, and Zhongfei Zhang.
\newblock Multi-content interaction network for few-shot segmentation.
\newblock \emph{ACM Transactions on Multimedia Computing, Communications and Applications}, 20\penalty0 (6):\penalty0 1--20, 2024{\natexlab{b}}.

\bibitem[Chen et~al.(2023{\natexlab{b}})Chen, Han, Gong, Bai, Ling, Luo, Chen, Ma, Zhang, Su, et~al.]{fengwu}
Kang Chen, Tao Han, Junchao Gong, Lei Bai, Fenghua Ling, Jing-Jia Luo, Xi Chen, Leiming Ma, Tianning Zhang, Rui Su, et~al.
\newblock Fengwu: Pushing the skillful global medium-range weather forecast beyond 10 days lead.
\newblock \emph{arXiv:2304.02948}, 2023{\natexlab{b}}.

\bibitem[Chen et~al.(2023{\natexlab{c}})Chen, Zhong, Zhang, Cheng, Xu, Qi, and Li]{fuxi}
Lei Chen, Xiaohui Zhong, Feng Zhang, Yuan Cheng, Yinghui Xu, Yuan Qi, and Hao Li.
\newblock Fuxi: A cascade machine learning forecasting system for 15-day global weather forecast.
\newblock \emph{npj Climate and Atmospheric Science}, 6\penalty0 (1):\penalty0 190, 2023{\natexlab{c}}.

\bibitem[Chen et~al.(2023{\natexlab{d}})Chen, Zhou, Du, Huang, Laudon, Chen, and Cui]{lifelongmoe}
Wuyang Chen, Yanqi Zhou, Nan Du, Yanping Huang, James Laudon, Zhifeng Chen, and Claire Cui.
\newblock Lifelong language pretraining with distribution-specialized experts.
\newblock In \emph{International Conference on Machine Learning}, pages 5383--5395. PMLR, 2023{\natexlab{d}}.

\bibitem[Dao et~al.(2022)Dao, Fu, Ermon, Rudra, and R{\'e}]{flashattention}
Tri Dao, Dan Fu, Stefano Ermon, Atri Rudra, and Christopher R{\'e}.
\newblock Flashattention: Fast and memory-efficient exact attention with io-awareness.
\newblock In \emph{Advances in Neural Information Processing Systems}, pages 16344--16359, 2022.

\bibitem[Dosovitskiy(2020)]{vit}
Alexey Dosovitskiy.
\newblock An image is worth 16x16 words: Transformers for image recognition at scale.
\newblock \emph{arXiv:2010.11929}, 2020.

\bibitem[Gan et~al.(2024)Gan, Man, Zhang, and Shao]{ewmoe}
Lihao Gan, Xin Man, Chenghong Zhang, and Jie Shao.
\newblock Ewmoe: An effective model for global weather forecasting with mixture-of-experts.
\newblock \emph{arXiv:2405.06004}, 2024.

\bibitem[Gao et~al.(2025)Gao, Wu, Shu, Dong, Xu, Chen, Yan, Wen, Hu, Wang, et~al.]{oneforecast}
Yuan Gao, Hao Wu, Ruiqi Shu, Huanshuo Dong, Fan Xu, Rui Chen, Yibo Yan, Qingsong Wen, Xuming Hu, Kun Wang, et~al.
\newblock Oneforecast: A universal framework for global and regional weather forecasting.
\newblock \emph{arXiv:2502.00338}, 2025.

\bibitem[Goswami et~al.(2024)Goswami, Soutif-Cormerais, Liu, Kamath, Twardowski, van~de Weijer, et~al.]{goswami2024resurrecting}
Dipam Goswami, Albin Soutif-Cormerais, Yuyang Liu, Sandesh Kamath, Bart Twardowski, Joost van~de Weijer, et~al.
\newblock Resurrecting old classes with new data for exemplar-free continual learning.
\newblock In \emph{Proceedings of the IEEE Conference on Computer Vision and Pattern Recognition}, pages 28525--28534, 2024.

\bibitem[He et~al.(2016)He, Zhang, Ren, and Sun]{he2016deep}
Kaiming He, Xiangyu Zhang, Shaoqing Ren, and Jian Sun.
\newblock Deep residual learning for image recognition.
\newblock In \emph{Proceedings of the IEEE Conference on Computer Vision and Pattern Recognition}, pages 770--778, 2016.

\bibitem[Hersbach et~al.(2020)Hersbach, Bell, Berrisford, Hirahara, Hor{\'a}nyi, Mu{\~n}oz-Sabater, Nicolas, Peubey, Radu, Schepers, et~al.]{era5}
Hans Hersbach, Bill Bell, Paul Berrisford, Shoji Hirahara, Andr{\'a}s Hor{\'a}nyi, Joaqu{\'\i}n Mu{\~n}oz-Sabater, Julien Nicolas, Carole Peubey, Raluca Radu, Dinand Schepers, et~al.
\newblock The era5 global reanalysis.
\newblock \emph{Quarterly Journal of the Royal Meteorological Society}, 146\penalty0 (730):\penalty0 1999--2049, 2020.

\bibitem[Jacobs et~al.(1991)Jacobs, Jordan, Nowlan, and Hinton]{jacobs1991adaptive}
Robert~A Jacobs, Michael~I Jordan, Steven~J Nowlan, and Geoffrey~E Hinton.
\newblock Adaptive mixtures of local experts.
\newblock \emph{Neural computation}, 3\penalty0 (1):\penalty0 79--87, 1991.

\bibitem[Keisler(2022)]{keisler}
Ryan Keisler.
\newblock Forecasting global weather with graph neural networks.
\newblock \emph{arXiv:2202.07575}, 2022.

\bibitem[Kim et~al.(2024)Kim, Yu, and Hwang]{Kim_2024_CVPR}
Beomyoung Kim, Joonsang Yu, and Sung~Ju Hwang.
\newblock Eclipse: Efficient continual learning in panoptic segmentation with visual prompt tuning.
\newblock In \emph{Proceedings of the IEEE Conference on Computer Vision and Pattern Recognition}, pages 3346--3356, 2024.

\bibitem[Kurth et~al.(2023)Kurth, Subramanian, Harrington, Pathak, Mardani, Hall, Miele, Kashinath, and Anandkumar]{fourcastnet}
Thorsten Kurth, Shashank Subramanian, Peter Harrington, Jaideep Pathak, Morteza Mardani, David Hall, Andrea Miele, Karthik Kashinath, and Anima Anandkumar.
\newblock Fourcastnet: Accelerating global high-resolution weather forecasting using adaptive fourier neural operators.
\newblock In \emph{Proceedings of the Platform for Advanced Scientific Computing Conference}, 2023.

\bibitem[Lam et~al.(2023)Lam, Sanchez-Gonzalez, Willson, Wirnsberger, Fortunato, Alet, Ravuri, Ewalds, Eaton-Rosen, Hu, et~al.]{graphcast}
Remi Lam, Alvaro Sanchez-Gonzalez, Matthew Willson, Peter Wirnsberger, Meire Fortunato, Ferran Alet, Suman Ravuri, Timo Ewalds, Zach Eaton-Rosen, Weihua Hu, et~al.
\newblock Learning skillful medium-range global weather forecasting.
\newblock \emph{Science}, 382\penalty0 (6677):\penalty0 1416--1421, 2023.

\bibitem[Li et~al.(2020)Li, Kovachki, Azizzadenesheli, Liu, Bhattacharya, Stuart, and Anandkumar]{li2020fno}
Zongyi Li, Nikola Kovachki, Kamyar Azizzadenesheli, Burigede Liu, Kaushik Bhattacharya, Andrew Stuart, and Anima Anandkumar.
\newblock Fourier neural operator for parametric partial differential equations.
\newblock \emph{arXiv:2010.08895}, 2020.

\bibitem[Liu et~al.(2021)Liu, Lin, Cao, Hu, Wei, Zhang, Lin, and Guo]{liu2021swin}
Ze Liu, Yutong Lin, Yue Cao, Han Hu, Yixuan Wei, Zheng Zhang, Stephen Lin, and Baining Guo.
\newblock Swin transformer: Hierarchical vision transformer using shifted windows.
\newblock In \emph{Proceedings of the IEEE International Conference on Computer Vision}, pages 10012--10022, 2021.

\bibitem[Molteni et~al.(1996)Molteni, Buizza, Palmer, and Petroliagis]{molteni1996ecmwf}
Franco Molteni, Roberto Buizza, Tim~N Palmer, and Thomas Petroliagis.
\newblock The ecmwf ensemble prediction system: Methodology and validation.
\newblock \emph{Quarterly journal of the royal meteorological society}, 122\penalty0 (529):\penalty0 73--119, 1996.

\bibitem[Nguyen et~al.(2023)Nguyen, Brandstetter, Kapoor, Gupta, and Grover]{climax}
Tung Nguyen, Johannes Brandstetter, Ashish Kapoor, Jayesh~K. Gupta, and Aditya Grover.
\newblock Climax: A foundation model for weather and climate.
\newblock In \emph{International Conference on Machine Learning}, 2023.

\bibitem[Nguyen et~al.(2025)Nguyen, Shah, Bansal, Arcomano, Maulik, Kotamarthi, Foster, Madireddy, and Grover]{stormer}
Tung Nguyen, Rohan Shah, Hritik Bansal, Troy Arcomano, Romit Maulik, Rao Kotamarthi, Ian Foster, Sandeep Madireddy, and Aditya Grover.
\newblock Scaling transformer neural networks for skillful and reliable medium-range weather forecasting.
\newblock In \emph{Advances in Neural Information Processing Systems}, pages 68740--68771, 2025.

\bibitem[Pan et~al.(2024)Pan, Zhou, Cao, and Zha]{pan2024adaptive}
Youqi Pan, Wugen Zhou, Yingdian Cao, and Hongbin Zha.
\newblock Adaptive vio: Deep visual-inertial odometry with online continual learning.
\newblock \emph{arXiv:2405.16754}, 2024.

\bibitem[Park et~al.(2024)Park, Go, Kim, Woo, Ham, and Kim]{dit-moe}
Byeongjun Park, Hyojun Go, Jin-Young Kim, Sangmin Woo, Seokil Ham, and Changick Kim.
\newblock Switch diffusion transformer: Synergizing denoising tasks with sparse mixture-of-experts.
\newblock In \emph{European Conference on Computer Vision}, 2024.

\bibitem[Price et~al.(2025)Price, Sanchez-Gonzalez, Alet, Andersson, El-Kadi, Masters, Ewalds, Stott, Mohamed, Battaglia, et~al.]{gencast}
Ilan Price, Alvaro Sanchez-Gonzalez, Ferran Alet, Tom~R Andersson, Andrew El-Kadi, Dominic Masters, Timo Ewalds, Jacklynn Stott, Shakir Mohamed, Peter Battaglia, et~al.
\newblock Probabilistic weather forecasting with machine learning.
\newblock \emph{Nature}, 637\penalty0 (8044):\penalty0 84--90, 2025.

\bibitem[Richardson(2024)]{ifs}
D Richardson.
\newblock The thorpex interactive grand global ensemble (tigge).
\newblock In \emph{Geophysical Research Abstracts}, page 02815, 2024.

\bibitem[Ritchie et~al.(1995)Ritchie, Temperton, Simmons, Hortal, Davies, Dent, and Hamrud]{ritchie1995implementation}
Harold Ritchie, Clive Temperton, Adrian Simmons, Mariano Hortal, Terry Davies, David Dent, and Mats Hamrud.
\newblock Implementation of the semi-lagrangian method in a high-resolution version of the ecmwf forecast model.
\newblock \emph{Monthly Weather Review}, 123\penalty0 (2):\penalty0 489--514, 1995.

\bibitem[Seo et~al.(2024)Seo, Koh, Jeung, Lee, Kim, Lee, Cho, Choi, Kim, and Choi]{seo2024learning}
Minhyuk Seo, Hyunseo Koh, Wonje Jeung, Minjae Lee, San Kim, Hankook Lee, Sungjun Cho, Sungik Choi, Hyunwoo Kim, and Jonghyun Choi.
\newblock Learning equi-angular representations for online continual learning.
\newblock In \emph{Proceedings of the IEEE Conference on Computer Vision and Pattern Recognition}, pages 23933--23942, 2024.

\bibitem[Tan et~al.(2024)Tan, Zhou, Xiang, Wang, Wu, and Li]{tan2024semantically}
Yuwen Tan, Qinhao Zhou, Xiang Xiang, Ke Wang, Yuchuan Wu, and Yongbin Li.
\newblock Semantically-shifted incremental adapter-tuning is a continual vitransformer.
\newblock In \emph{Proceedings of the IEEE Conference on Computer Vision and Pattern Recognition}, pages 23252--23262, 2024.

\bibitem[Verma et~al.(2024)Verma, Heinonen, and Garg]{climode}
Yogesh Verma, Markus Heinonen, and Vikas Garg.
\newblock Clim{ODE}: Climate forecasting with physics-informed neural {ODE}s.
\newblock In \emph{International Conference on Learning Representations}, 2024.

\bibitem[Wu et~al.(2024)Wu, Hu, Wang, Pang, and Soricut]{wu2024omni}
Jialin Wu, Xia Hu, Yaqing Wang, Bo Pang, and Radu Soricut.
\newblock Omni-smola: Boosting generalist multimodal models with soft mixture of low-rank experts.
\newblock In \emph{Proceedings of the IEEE/CVF Conference on Computer Vision and Pattern Recognition}, pages 14205--14215, 2024.

\bibitem[Xiong et~al.(2023)Xiong, Ma, Huang, Zhang, Sun, and Tian]{xiong2023koopmanlab}
Wei Xiong, Muyuan Ma, Xiaomeng Huang, Ziyang Zhang, Pei Sun, and Yang Tian.
\newblock Koopmanlab: machine learning for solving complex physics equations.
\newblock \emph{APL Machine Learning}, 1\penalty0 (3), 2023.

\bibitem[Xu et~al.(2024)Xu, Ling, Zhang, Han, Chen, Ouyang, and Bai]{weathergft}
Wanghan Xu, Fenghua Ling, Wenlong Zhang, Tao Han, Hao Chen, Wanli Ouyang, and Lei Bai.
\newblock Generalizing weather forecast to fine-grained temporal scales via physics-ai hybrid modeling.
\newblock In \emph{Advances in Neural Information Processing Systems}, 2024.

\bibitem[Yan et~al.(2024)Yan, Wang, Ma, and Zhong]{yan2024orchestrate}
Hongwei Yan, Liyuan Wang, Kaisheng Ma, and Yi Zhong.
\newblock Orchestrate latent expertise: Advancing online continual learning with multi-level supervision and reverse self-distillation.
\newblock In \emph{Proceedings of the IEEE Conference on Computer Vision and Pattern Recognition}, pages 23670--23680, 2024.

\bibitem[Yang et~al.(2024)Yang, Jiang, Hou, Zhang, Chen, and Li]{yang2024multi}
Yuqi Yang, Peng-Tao Jiang, Qibin Hou, Hao Zhang, Jinwei Chen, and Bo Li.
\newblock Multi-task dense prediction via mixture of low-rank experts.
\newblock In \emph{Proceedings of the IEEE/CVF Conference on Computer Vision and Pattern Recognition}, pages 27927--27937, 2024.

\bibitem[Ye and Bors(2024)]{ye2024online}
Fei Ye and Adrian~G Bors.
\newblock Online task-free continual generative and discriminative learning via dynamic cluster memory.
\newblock In \emph{Proceedings of the IEEE Conference on Computer Vision and Pattern Recognition}, pages 26202--26212, 2024.

\bibitem[Yoon et~al.(2024)Yoon, Seo, Kim, Choi, and Cho]{probabilistic}
Donggeun Yoon, Minseok Seo, Doyi Kim, Yeji Choi, and Donghyeon Cho.
\newblock Probabilistic weather forecasting with deterministic guidance-based diffusion model.
\newblock In \emph{European Conference on Computer Vision}, pages 108--124, 2024.

\bibitem[Yu et~al.(2024{\natexlab{a}})Yu, Li, Ye, Zhang, Luo, Dai, Wang, and Chen]{diffcast}
Demin Yu, Xutao Li, Yunming Ye, Baoquan Zhang, Chuyao Luo, Kuai Dai, Rui Wang, and Xunlai Chen.
\newblock Diffcast: A unified framework via residual diffusion for precipitation nowcasting.
\newblock In \emph{Proceedings of the IEEE Conference on Computer Vision and Pattern Recognition}, pages 27758--27767, 2024{\natexlab{a}}.

\bibitem[Yu et~al.(2024{\natexlab{b}})Yu, Zhuge, Zhang, Hu, Wang, Lu, and He]{moeadapter}
Jiazuo Yu, Yunzhi Zhuge, Lu Zhang, Ping Hu, Dong Wang, Huchuan Lu, and You He.
\newblock Boosting continual learning of vision-language models via mixture-of-experts adapters.
\newblock In \emph{Proceedings of the IEEE/CVF Conference on Computer Vision and Pattern Recognition}, pages 23219--23230, 2024{\natexlab{b}}.

\bibitem[Zhang et~al.(2023)Zhang, Long, Chen, Xing, Jin, Jordan, and Wang]{nowcastnet}
Yuchen Zhang, Mingsheng Long, Kaiyuan Chen, Lanxiang Xing, Ronghua Jin, Michael~I Jordan, and Jianmin Wang.
\newblock Skilful nowcasting of extreme precipitation with nowcastnet.
\newblock \emph{Nature}, 619\penalty0 (7970):\penalty0 526--532, 2023.

\bibitem[Zhu et~al.(2024)Zhu, Cheng, Zhang, Liu, and Zhang]{zhu2024rcl}
Fei Zhu, Zhen Cheng, Xu-Yao Zhang, Cheng-Lin Liu, and Zhaoxiang Zhang.
\newblock Rcl: Reliable continual learning for unified failure detection.
\newblock In \emph{Proceedings of the IEEE Conference on Computer Vision and Pattern Recognition}, pages 12140--12150, 2024.

\end{thebibliography}
}

\end{document}